\documentclass[singlecolumn,letterpaper,12pt]{article}
\pdfoutput=1
\usepackage{graphicx}
\usepackage{bm}
\usepackage{graphicx}
\usepackage{latexsym}
\usepackage{epstopdf}
\usepackage{amsmath}
\usepackage{amsthm} 
\usepackage{times}

\usepackage{txfonts}
\usepackage{stfloats}
\usepackage{algorithm}
\usepackage{algorithmic}

\begin{document}

\date{}

\title{Renewable Energy Prediction using Weather Forecasts for Optimal Scheduling in HPC Systems}
\author{
Ankur Sahai \\
Johannes Gutenberg University of Mainz, Germany
}

\hyphenpenalty=1000
\setlength{\parindent}{10pt}
\setlength{\parskip}{1ex} 
\maketitle
\thispagestyle{empty}
\noindent
\begin{abstract}
The objective of the GreenPAD project is to use green energy to power data-centers. As a part of this it is important to predict the Wind energy for efficient scheduling (executing jobs that require higher energy when there is more green energy available and vice-versa). For predicting the wind energy we first analyze the historical data to find a statistical model that gives relation between wind energy and weather attributes. Then we use this model on the weather forecast data to predict the green energy availability in the future.
\end{abstract}

\section{Introduction}
\label{sec:introduction}
As a part of the GreenPAD project we analyze the wind energy (E.ON) and weather data (DWD) using simple statistical techniques such as correlation and using Matlab for detailed analysis. Objective of GreenPAD project is to maximize the utilization of green energy (wind, solar and biomass) to power the green data centers.

This is a challenge as the energy that can be transmitted to the main grid is restricted by the capacity of the transformers. When the combination of the wind energy and conventional energy exceeds the transformer capacity it has to be used locally in order to avoid wastage. This is addressed by running a green data center whose energy consumption can be controlled as per the green energy availability; as it is not possible to control the energy consumption pattern of the households.

\begin{figure}
\centering
\includegraphics[width=125mm]{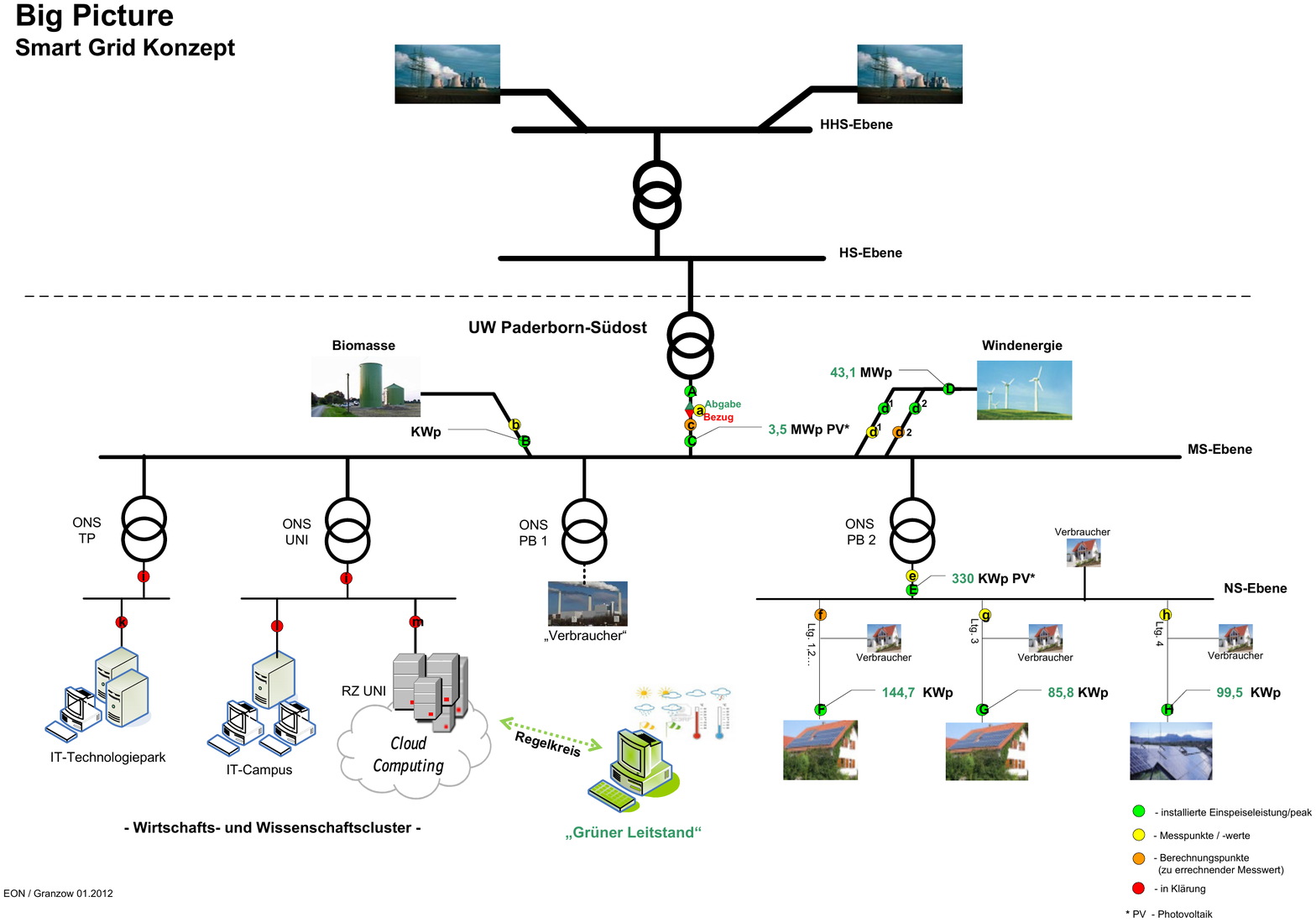}
\caption{GreenPAD model}
\label{model}
\end{figure}

Specifically we focus on wind energy for an analysis similar to paper \cite{solar-energy-predicition}. We start with simple techniques such as finding the correlation between wind energy and weather attributes and use Matlab plots for fine grained analysis. Next we try to find an equation that relates wind energy to weather attributes. This is useful for the cases where there is one or more parameters missing e.g. no wind energy data but weather attributes. The schema of GreenPAD is depicted in Figure \ref{model} .

\section{Motivation}
\label{sec:motivation}

We try to use the weather forecast to predict the wind energy by finding the exact relation between wind energy and weather attributes. From the perspective of GreenPAD project, wind energy prediction is critical for scheduling the jobs on the \emph{green data center}. For example, jobs that have higher energy requirement could be scheduled at times when there is more green energy available and vice-versa.

\section{Analysis of Wind energy and Weather data}
\label{sec:wind-energy-weather-analysis}

\begin{figure}
\centering
\hspace*{-.5cm}\includegraphics[height=110mm]{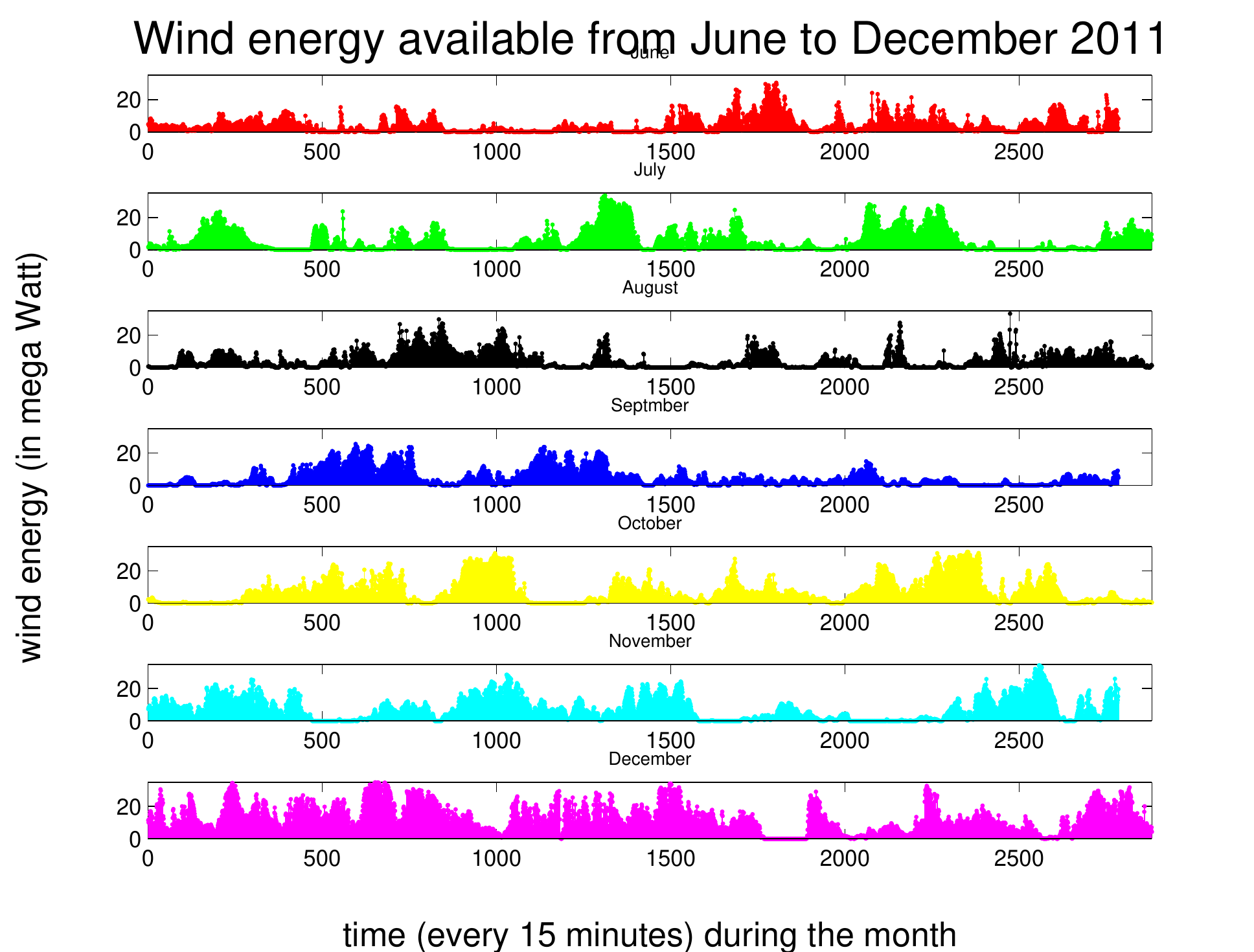}
\label{wind-energy-D}
\caption{Wind energy expenditure measured at point D in Figure \ref{model}}
\end{figure}

\begin{figure}
\centering
\hspace*{-2cm}\includegraphics[width=150mm]{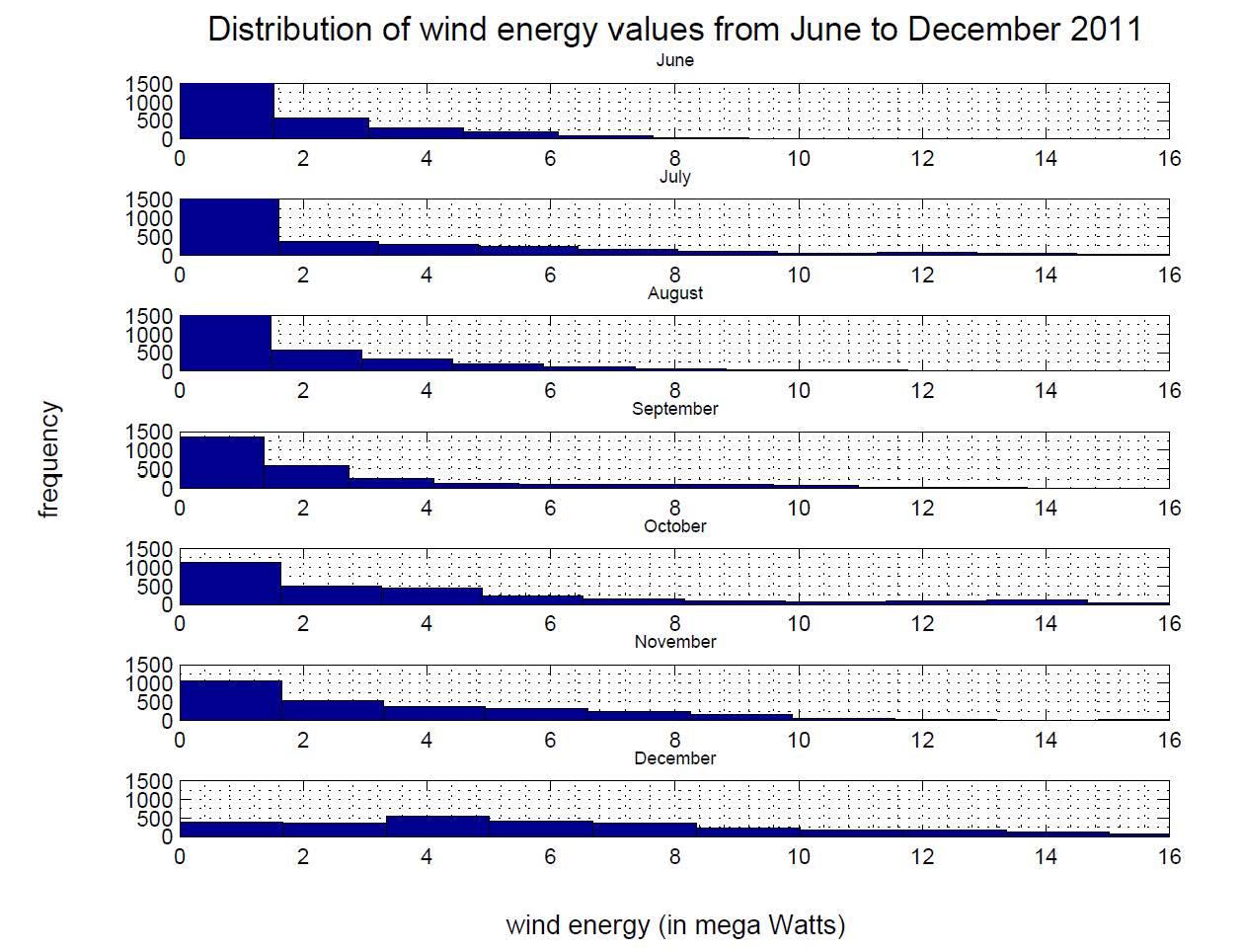}
\label{wind-energy-hist}
\caption{Distribution of wind energy intensities across months}
\end{figure}

November and December show high wind energy availability (which depends upon wind speed)  followed by July as seen from Figure \ref{wind-energy-D}. During December the wind energy intensities were uniformly distributed i.e. high and low wind speeds were almost equally likely whereas August showed the least variation with wind energy availability being very low (0-1 MW) most of the times and hardly exceeding 8 W.

\begin{table}
\begin{center}
\begin{tabular}{ | l | l | l | l | l | l | l | l |l |}
\hline
Covariance & WE & Temp & CC & AP & WS & WD & PP & SS \\ \hline
Wind Energy & 1.00 & -0.10 & 0.11 & 0.01 & 0.68 & 0.28 & 0.11  & -0.09\\ \hline
Temperature & -0.10 & 1.00 & 0.05 & 0.02 & 0.08 & 0.16 & -0.02 & 0.20 \\ \hline
Cloud cover & 0.11 & 0.05 & 1.00 & -0.36 & 0.17 & 0.30 & -0.01 & -0.19 \\ \hline
Air Pressure & 0.01 & 0.02 & -0.36 & 1.00 & 0.02 & 0.03 & 0.14 & 0.00\\ \hline
Wind speed & 0.68 & 0.08 & 0.17 & 0.02 & 1.00 & 0.34 & 0.09 & 0.03\\ \hline
Wind direction & 0.28 & 0.16 & 0.30 & 0.03 & 0.34 & 1.00 & 0.08 & -0.04 \\ \hline
Precipitation & 0.11 & -0.02 & -0.01 & 0.14 & 0.09 & 0.08 & 1.00 & -0.04 \\ \hline
SunShine & -0.09 & 0.20 & -0.19 & 0.31 & 0.03 & -0.04 & -0.04 & 1.00 \\ \hline
\end{tabular}
\caption{Correlation between wind energy and weather attributes}
\label{table1}
\end{center}
\end{table}

\begin{figure}
\centering
\includegraphics[width=100mm,height=75mm]{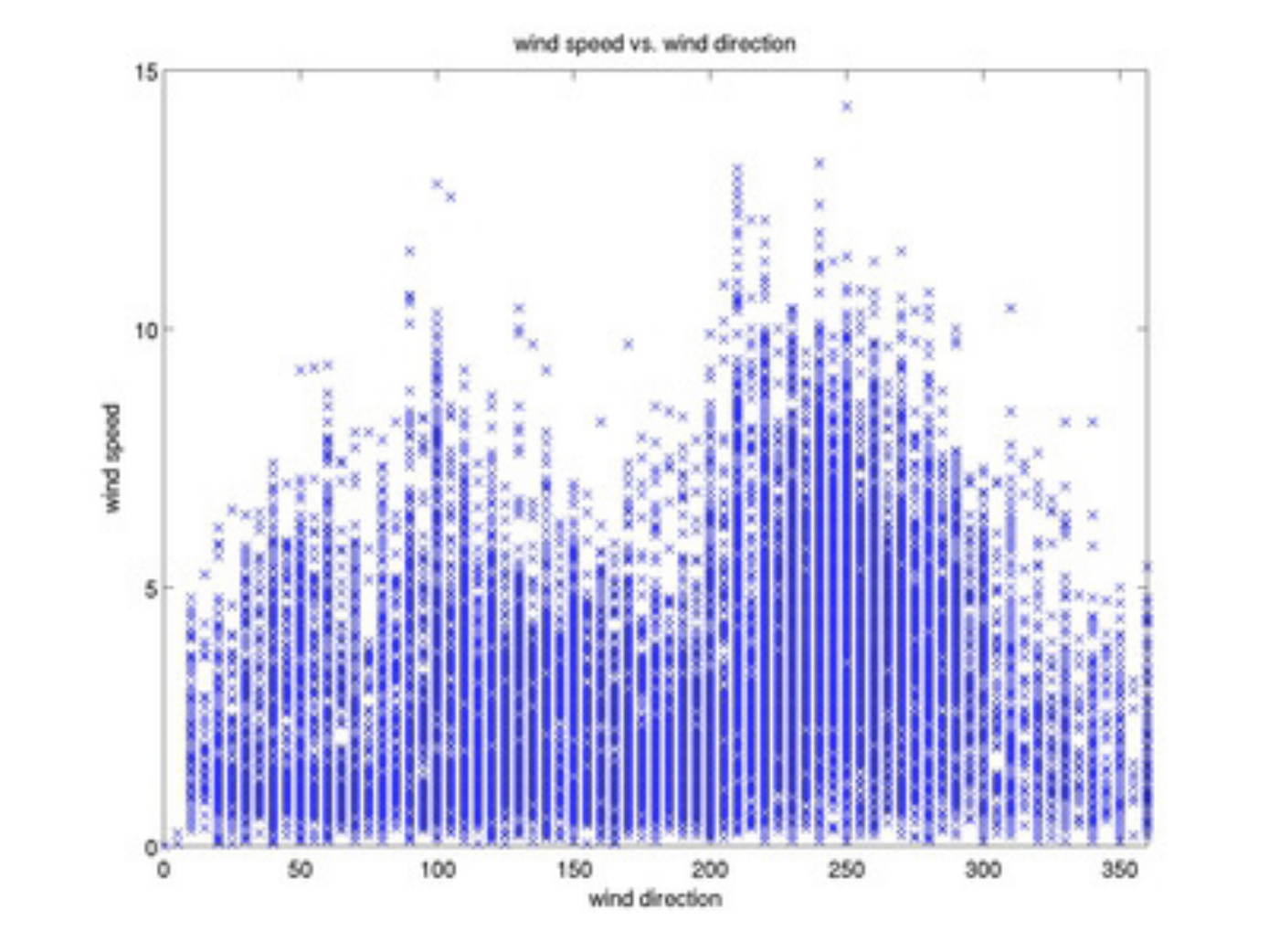}
\caption{Plot between wind direction and wind speed}
\label{wind-dir-speed}
\end{figure}

In Table \ref{table1} wind energy shows a high degree of correlation with wind speed (0.68) and wind direction (0.31). This is followed by Air pressure (0.17) and Precipitation (0.11).  Temperature and Sunshine show negative correlation. Negative correlation between solar and wind energy of -.948.

High correlation between wind speed and wind direction in Table \ref{table1} is explained by Figure \ref{wind-dir-speed} where the winds coming from east (90 degrees) and west (270 degrees) have higher speeds which may corresponds to the high-pressure areas due to low temperatures and winds coming coming from the ocean respectively.

\begin{figure}
\centering
\includegraphics[width=100mm,height=75mm]{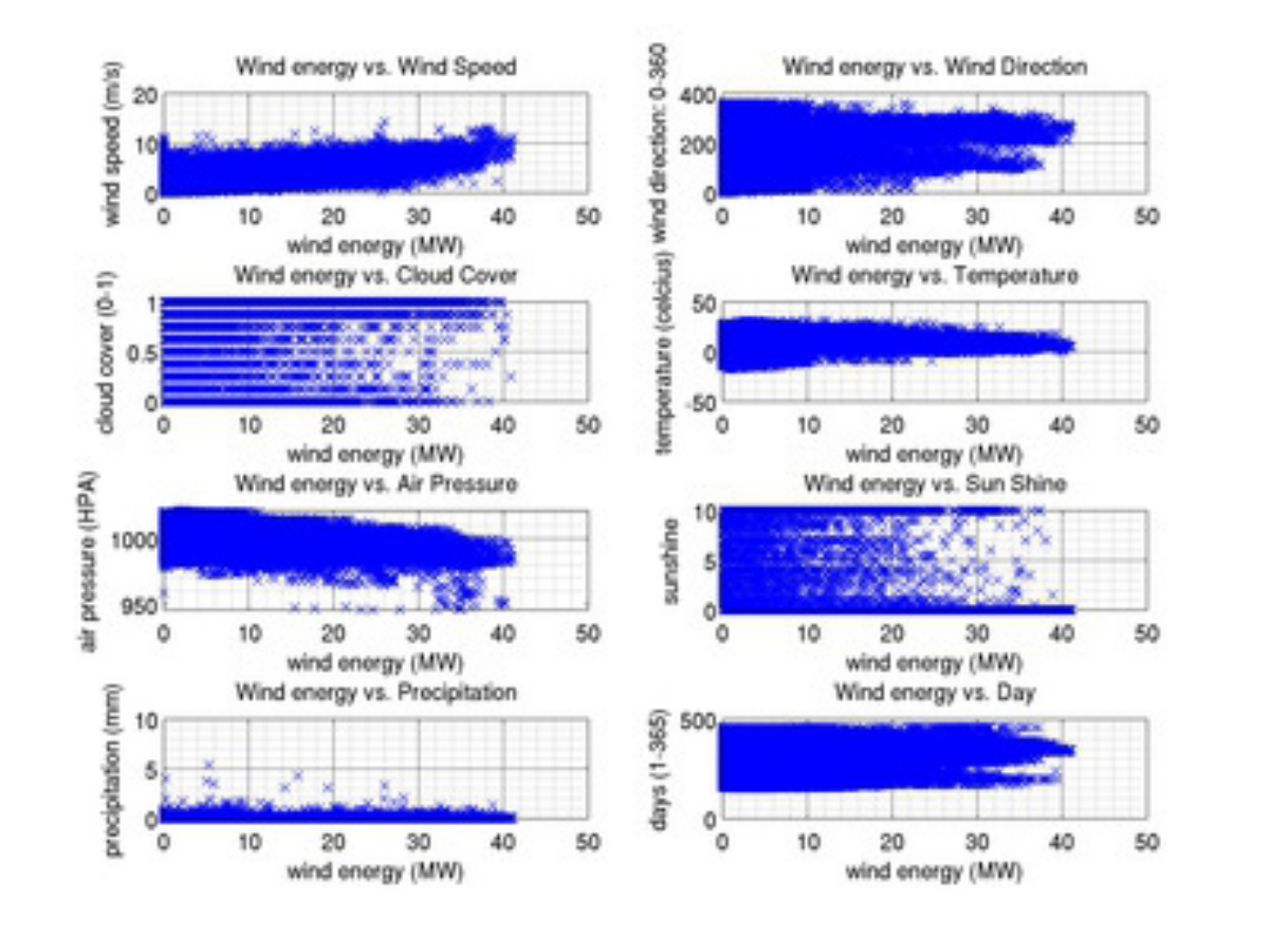}
\caption{Relationship of wind energy with weather attributes}
\label{wind-energy}
\end{figure}

\begin{figure}
\centering
\includegraphics[width=100mm,height=60mm]{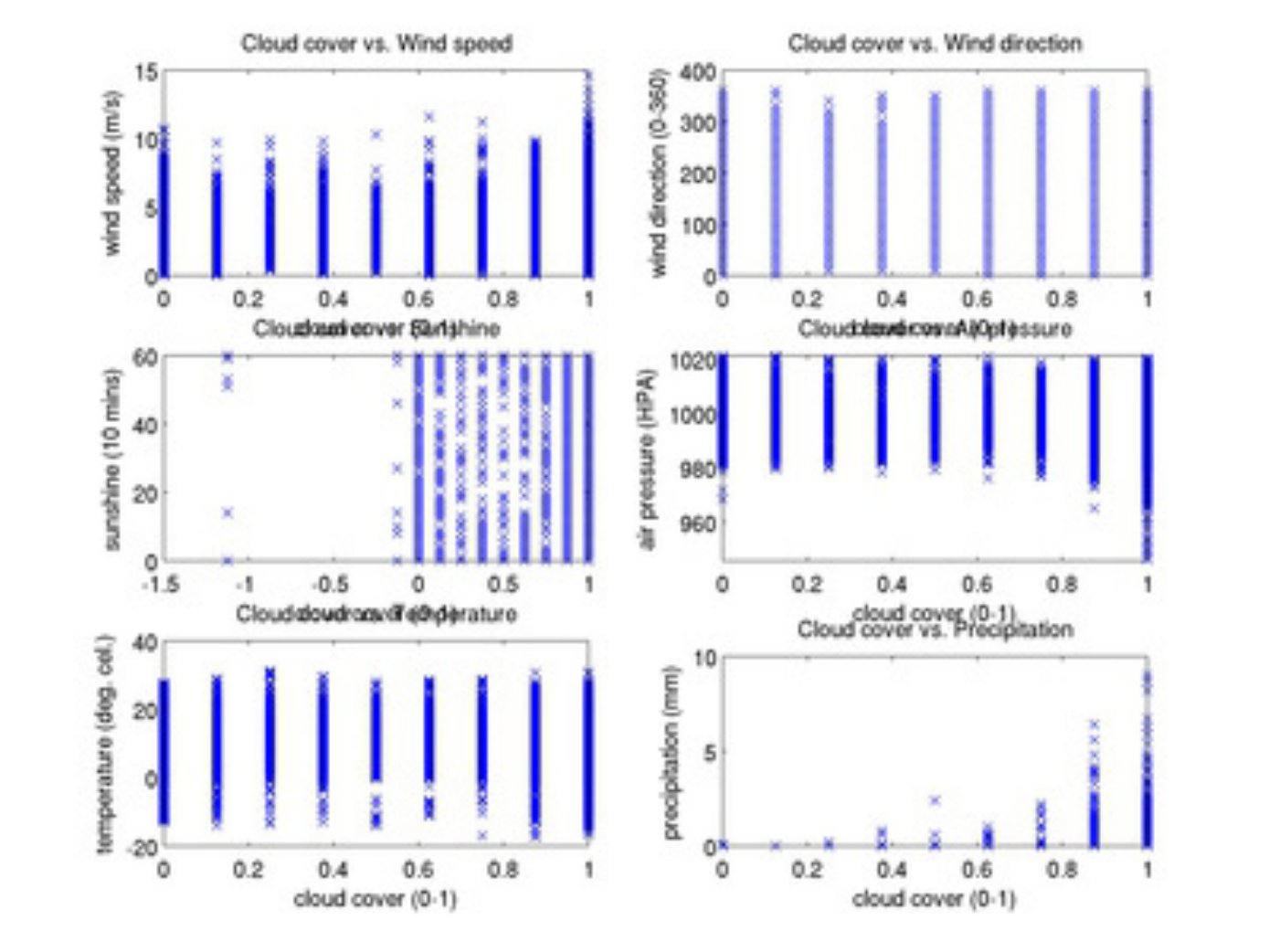}
\caption{Relationship of wind speed (and other attributes) with cloud cover}
\label{wind-speed-cloud-cover}
\end{figure}

\begin{figure}
\centering
\includegraphics[width=100mm,height=60mm]{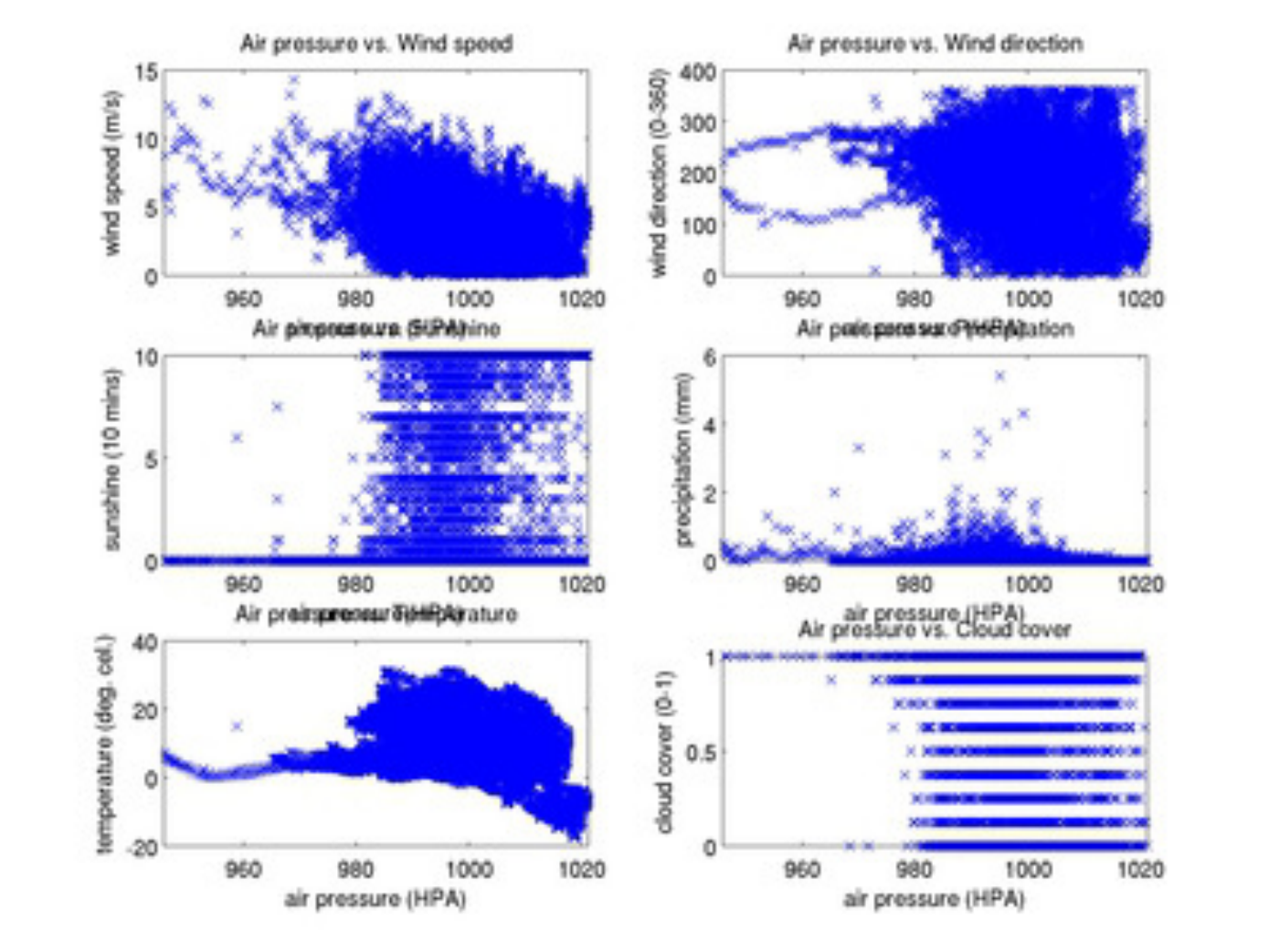}
\caption{Relationship of wind speed (and other attributes) with air pressure}
\label{wind-speed-air-pressure}
\end{figure}

In Figure \ref{wind-energy} we find wind speed and Wind direction to have the highest correlation with the wind energy. Although the correlation between wind energy and wind speed is obvious we attribute the high correlation with wind direction as high speed winds usually come from specific directions e.g. east to west may be due to winds coming from ocean or formation of other low pressure areas. In the plot between wind energy and days winters show higher wind intensity. In the plot with wind direction the spikes at around 90 and 270 degrees is because the winds coming from east and west have relatively higher intensities when compared to other directions. 

In the plot with cloud cover there is higher wind speed at times when there is very low or high cloud cover which corresponds to clear sky and intense clouds. In the next plot with sunshine again higher wind speed is observer when there is either higher or lower sunshine. Whereas low sunshine periods correspond to winters, higher sunshine would create a low-pressure area thus increasing the winds blowing inwards towards this area. In the next plot there is higher wind speed when the temperature ranges between around between 0 and 20 degree celsius as compared to other times. Wind speed increases with air pressure as usually higher air pressure seems to resist wind flow. Precipitation shows little correlation with wind speed. 

Cloud cover shows higher correlation with wind speed (Figure \ref{wind-speed-cloud-cover}) with higher values of cloud cover corresponding to higher wind speeds. However, higher wind speeds are not observed for partially cloudy condition(cloud cover value of around 0.5). Air pressure (Figure \ref{wind-speed-air-pressure}) only shows higher correlation with wind speed at lower air pressure values (940-980 HPA) where it is inversely related (wind speeds increase with decrease in air pressure).

\begin{figure}
\centering
\includegraphics[width=100mm,height=75mm]{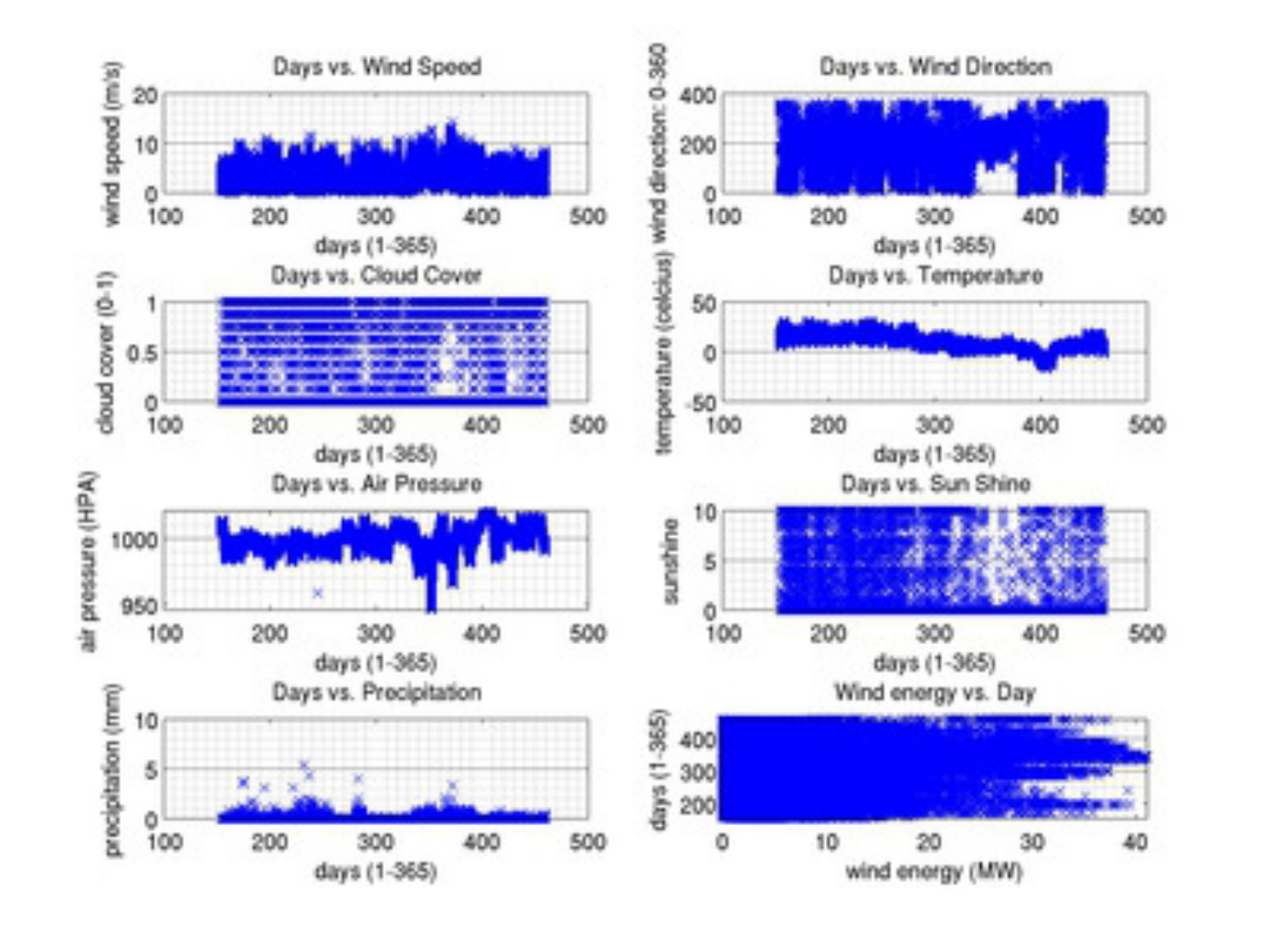}
\caption{Plot between days of year (mod 365) with weather attributes and wind energy}
\label{days}
\end{figure}

In Figure \ref{days} (days on x-axis have to interpreted as days mod 365), higher wind energy is observer during December and January. Wind directions seems to shift at regular intervals thus creating a zig-zag shape except during winters where the winds blow exclusively from west possibly due to creation of a low-pressure in the east. Temperature as expected increases during summers and decreases during winters. Sunshine seems to be uniform apart from the period around December where it is educed due to cloudy conditions. Wind energy peaks during winters which has a direct correlation with wind speed. Precipitation as expected is higher during the rainy seasons during August. Air pressure slowly decreases until winters and then increases. Cloud cover remains uniform throughout year apart from period during February.

\section{Statistical Model for Wind Energy Prediction using Weather data}
\label{sec:statistical-prediction-model}
The equation for relation of wind energy with weather attributes (using correlation regression) is as follows:

\begin{eqnarray}
Predicted Wind energy &= - .84 *Temperature -  .96*CloudCover \nonumber \\
& - .89*AirPressure + .71*WindSpeed \nonumber \\ 
& - .15*WindDirection - .78*Precipitation \nonumber \\ 
& - 1.02*SunShine
\label{eqn1}
\end{eqnarray}

For deriving equation (\ref{eqn1}) we use correlation regression. Using \ref{table1} find equations that define relationship of one weather attribute with remaining attributes and wind energy. Second we solve these equations with wind energy as the right hand side.

\section{Survey of machine learning techniques}
\label{sec:machine-learning}

Some of the simple scheduling algorithms tried were: Greedy, Randomized-Greedy which were compared with Brute-force which checks the cost of all possible schedules and picks the one with lowest cost. However, it is to be noted that this is not practical as it has very high time-complexity.

\subsection{Binary decision tree}
We propose precomputing different job schedules and assigning jobs based on current weather using decision trees. Job schedules correspond to different weather conditions and we pick a schedule based on the current weather condition.

\subsection{Bayesian learning}
We propose recovering lost or missing data and predicting weather forecasts based on multi-variable Bayesian learning. In cases where there is weather or wind energy data missing in the dataset we use Bayesian networks implemented using tools like OpenBayes using bayesian classifiers to predict this information.

\section{Conclusion}
\label{sec:conclusion}
Hence we are able to predict green (wind) energy using weather forecasts using the statistical (regression) model obtained by analyzing the historical weather and energy data. Using the green energy prediction obtained from the statistical model we are able to precompute job schedules for maximizing the green energy utilization in the future.

We find a higher degree of correlation between wind energy and wind direction which indicates that the winds coming from certain directions - east and west had higher intensities. Maximum green energy is available during winters i.e. during the months of December and January.

\end{document}